# Adaptive Bidirectional Backpropagation: Towards Biologically Plausible Error Signal Transmission in Neural Networks

Hongyin Luo* MIT CSAIL hyluo@mit.edu

Jie Fu* National University of Singapore jie.fu@u.nus.edu

James Glass MIT CSAIL glass@mit.edu

**Abstract**

The back-propagation (BP) algorithm has been considered the de-facto method for training deep neural networks. It back-propagates errors from the output layer to the hidden layers in an exact manner using the transpose of the feedforward weights. However, it has been argued that this is not biologically plausible because back-propagating error signals with the exact incoming weights is not considered possible in biological neural systems. In this work, we propose a biologically plausible paradigm of neural architecture based on related literature in neuroscience and asymmetric BP-like methods. Specifically, we propose two bidirectional learning algorithms with trainable feedforward and feedback weights. The feedforward weights are used to relay activations from the inputs to target outputs. The feedback weights pass the error signals from the output layer to the hidden layers. Different from other asymmetric BP-like methods, the feedback weights are also plastic in our framework and are trained to approximate the forward activations. Preliminary results show that our models outperform other asymmetric BP-like methods on the MNIST and the CIFAR-10 datasets.

## 1 Introduction

Back-propagation (BP) algorithm is the combination of reverse-mode automatic differentiation [1] and steepest descent [10] which has been considered the de-facto method for training deep neural networks (DNNs). It back-propagates errors from output to input layer by layer in an exact manner. However, it has been argued that it is not biologically possible for learning in the brain to involve precise, symmetric backward channels [6, 15, 3, 12].

In the early days of deep learning, unsupervised pre-training with Boltzmann machines used to be applied before fine-tuning with BP [7], which does not involve symmetric weights and is biologically motivated. Recently, there has been a rising interest in developing both biologically feasible and practical alternatives for BP. In [22], target-propagation (TP)[12], whose objective is to let each layer to reproduce outputs the previous layer, is used to train a recurrent neural network for natural language processing tasks. The authors in [15] propose feedback-alignment (FA) model and showed that for BP-like methods, the weights used in the feedback pass do not have to be the transpose of feedforward weights. The direct feedback-alignment (DFA) model proposed in [17] suggest that error signals could be transmitted directly from output layer to any hidden layer with random and fixed matrices. One of the key requirements in FA and DFA model is that the feedback is random and fixed.

On the other hand, due to the literature in neuroscience, long-term potentiation (LTP) is considered an essential step in human memory and learning [16, 4]. As introduced in LTP, strong links between neurons are established starts from the neural adjustment step that one of the neurons moves more ion receptors onto the membrane of its dendrites. As a result, more ions can be captured, which amplifies the electrical impulses.

Based on the principles of LTP and the hypothesis that the feedback weights are plastic [2], we propose a more biological plausible perceptron paradigm and two bidirectional learning models. In the bidirectional learning models, the feedforward weights are adjusted in forward phase, and feedback weights are learned in backward phase. Our proposed models dispel the assumption that the feedback weights have to be random and fixed. The feedback weights are trained to approximate the forward activations during training. Experiments on benchmark datasets show that our models outperform the FA and DFA counterparts, which use fixed feedback weights to transmit error signals. We also provide preliminary analysis on why transmitting error signals with adaptive weights outperforms using fixed weights. To the best of our knowledge, this is the first research attempt to show that adaptive asymmetric feedback channels are more effective than random and fixed feedback channels in DNNs.

*Equal contribution

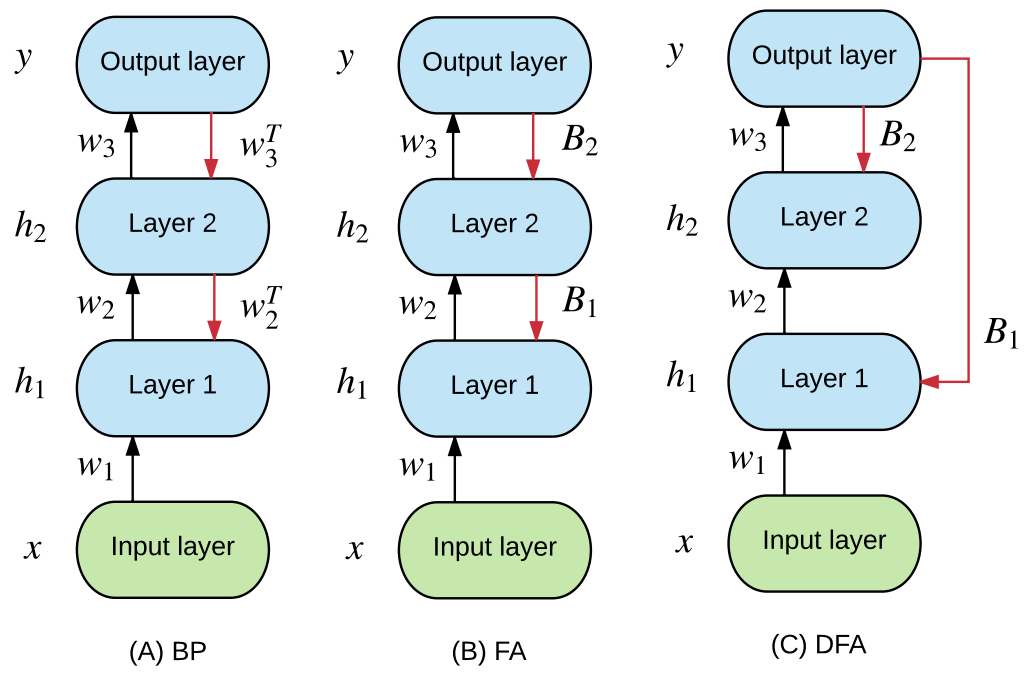

Figure 1: Back-propagation (BP), feedback-alignment (FA) and direct feedback-alignment (DFA) model. Black arrows represent forward activation paths. Red arrows indicate error (gradient) propagation paths. (Modified from [17]).

## 2 Background

Following the notation in [14], let $(x, y)$ be a mini-batch of input-output of size 1. The DNN we consider here has 2 hidden layers. $w_i$ are the weights (absorbing biases) connecting the previous layer to a unit in the $i$-th hidden layer. The activations can be computed as

$$\boldsymbol{a}_1 = \boldsymbol{w}_1\boldsymbol{x}, \boldsymbol{h}_1 = \sigma(\boldsymbol{a}_1) \tag{1}$$

$$\boldsymbol{a}_2 = \boldsymbol{w}_2\boldsymbol{h}_1, \boldsymbol{h}_2 = \sigma(\boldsymbol{a}_2) \tag{2}$$

$$\boldsymbol{a}_y = \boldsymbol{w}_3\boldsymbol{h}_2, \hat{\boldsymbol{y}} = \sigma_y(\boldsymbol{a}_y) \tag{3}$$

where $\sigma_y(\cdot)$ the activation function used in the output layer and $\sigma(\cdot)$ is activation function used in hidden layers.

The loss and the gradient at the output layer are:

$$\overrightarrow{l} = -\boldsymbol{y} \cdot \log\hat{\boldsymbol{y}} - (1 - \boldsymbol{y}) \cdot \log(1 - \hat{\boldsymbol{y}}) \tag{4}$$

$$\frac{\partial l}{\partial \boldsymbol{a}_y} = \hat{\boldsymbol{y}} - \boldsymbol{y} = \boldsymbol{e} \tag{5}$$

The gradients for hidden layers for the back-propagation (BP) are:

$$\delta\boldsymbol{a}_2 = \frac{\partial l}{\partial a_2} = (\boldsymbol{w}_3^T \delta a_y) \odot \sigma'(\boldsymbol{a}_2) \tag{6}$$

$$\delta\boldsymbol{a}_1 = \frac{\partial l}{\partial \boldsymbol{a}_1} = (\boldsymbol{w}_2^T \delta\boldsymbol{a}_2) \odot \sigma'(\boldsymbol{a}_1) \tag{7}$$

where $\sigma'(\cdot)$ is the derivative of the activation function and $\odot$ is an element-wise multiplication operator.

For feedback alignment (FA), as shown in Figure 1, the hidden layer update directions are:

$$\delta\boldsymbol{a}_2 = (\boldsymbol{B}_2\boldsymbol{e}) \odot \sigma'(\boldsymbol{a}_2) \tag{8}$$

$$\delta \boldsymbol{a}_1 = (\boldsymbol{B}_1 \delta \boldsymbol{a}_2) \odot \sigma'(\boldsymbol{a}_1) \tag{9}$$

In direct feedback alignment (DFA) model, error signals are transmitted directly from the output layer to each hidden layer. Gradients for layer 1 are calculated differently from FA, which is

$$\delta \boldsymbol{a}_1 = (\boldsymbol{B}_1 \boldsymbol{e}) \odot \sigma'(\boldsymbol{a}_1) \tag{10}$$

where $\boldsymbol{B}_i$ is a fixed random weighted matrix with appropriate dimension.

The weight updates for BP, FA and DFA methods are calculated as

$$\delta \boldsymbol{w}_1 = -\delta \boldsymbol{a}_1 \boldsymbol{x}^T, \delta \boldsymbol{w}_2 = -\delta \boldsymbol{a}_2 h_1^T, \delta \boldsymbol{w}_3 = -\boldsymbol{e}\boldsymbol{h}_2^T \tag{11}$$

where we ignore the learning rates, and the random weights are only used to transfer gradients back to hidden neurons. In BP, updating the individual weights needs to store the weights used in the forward pass.

Although FA and DFA with random and fixed feedback weights are more biologically plausible than BP, the feedback weights in the brain are plastic too [3]. It has been shown in [15] that the forward weights $\boldsymbol{w}_i$ used in FA learns to resemble the pseudo-inverse of the feedback random weights $\boldsymbol{B}_i$. Therefore, it would be desirable to prevent the forward weights $\boldsymbol{w}_i$ from becoming too similar to a random matrix. We will demonstrate that the model can be optimized by bidirectional training later.

The conventional DNNs with BP, FA, and DFA are unidirectional in the sense that they only learn how to map inputs to target outputs. In this paper, based on related literature in neuroscience, we propose a paradigm of biologically plausible perceptron model. Then we propose bidirectional feedback alignment (BFA) and bidirectional direct feedback alignment (BDFA) model, which connect neurons by two sets of trainable weights for the forward and the backward processes, respectively. A DNN with either BFA or BDFA is trained to predict outputs and generate feature maps simultaneously.

# 3 Biologically Plausible Perceptron

Classical perceptrons trained with gradient descent algorithms need to back-propagate error signals based on the exact feedforward synaptic weights, which is considered impossible in a biological neural system [20]. On the other hand, long-term potentiation (LTP) is considered an essential part of biological memory and learning in cognitive science [16, 4]. In this section, we first briefly describe the LTP mechanism and then propose a more biologically plausible perceptron paradigm.

## 3.1 Long-term Potentiation

Biological neurons are connected by synapse, including axons and dendrites, where axons emit signals, and dendrites of the next neuron receive the electrical impulses released by axons [8]. However, axons and dendrites are separated by synaptic clefts, and the axons send electrical impulses by releasing ions into synaptic cleft [5]. The ions are captured by receptors on the cell membrane of dendrites [13]. The architecture is shown in Figure 2.

When a synapse transmits neural signals from neuron $\mathcal{N}_1$ to neuron $\mathcal{N}_2$ and is repeatedly simulated, neuron $\mathcal{N}_2$ will release more receptors on its dendrites and thus capture more ions [4]. This procedure reduces the ion concentration of the synaptic cleft between $\mathcal{N}_1$ and $\mathcal{N}_2$, which encourages $\mathcal{N}_1$ to release more ions [4]. Thus, a stronger connection between neuron $\mathcal{N}_1$ and $\mathcal{N}_2$ is established due to the LTP procedure [4]. LTP adjusts links among neurons and plays a significant role in forming memory and learning [4].

## 3.2 Biologically Plausible Perceptron Model

The first step of synaptic adjustment between neuron $\mathcal{N}_1$ and $\mathcal{N}_2$ is that $\mathcal{N}_2$ adjusts the quantity of receptors on its dendrites, which is an important observation from LTP procedure. Based on this principle, we propose a more biologically plausible perceptron (BioPP) model.

The components of BioPP model are described as follow,

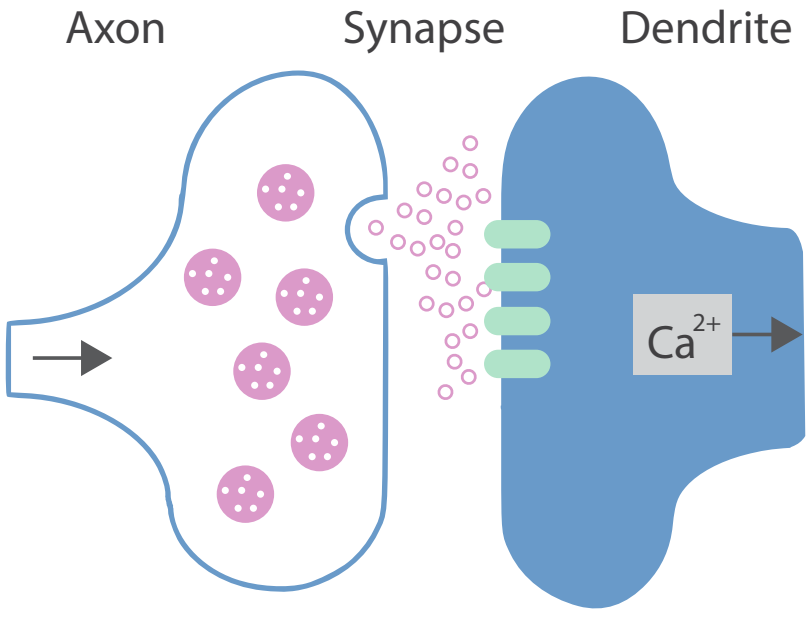

Figure 2: Architecture of neural connection between axon and dendrite. This image is reproduced from [21]

- **Signals** There are two sets of signals in BioPP architecture
  - Feedforward Signals: Signals propagated forward in network for inference
  - Error Signals: Signals propagated backward for adjusting synaptic weights
- **Weights** Weights stand for the quantity of signal a perceptron decides to capture from input or adjacent neurons. It should be noted that the quantity of the error signals taken by a perceptron is also decided by itself. A BioPP adjusts its own weights based on the incoming error signals, and then sends error signals to other neurons.
- **Activations and Biases** The definition of activations and biases follows that of the standard DNNs trained with BP.

The architecture of BioPP is shown in Figure 3, where the green circles are neurons, the blue curves stand for forward synaptic weights. The red curves stand for backward synaptic weights. The blue squares and the red squares are receptors for the forward and the backward synapses, respectively. It is worth noting that according to the definition of BioPP, weights are adjusted by the receptors.

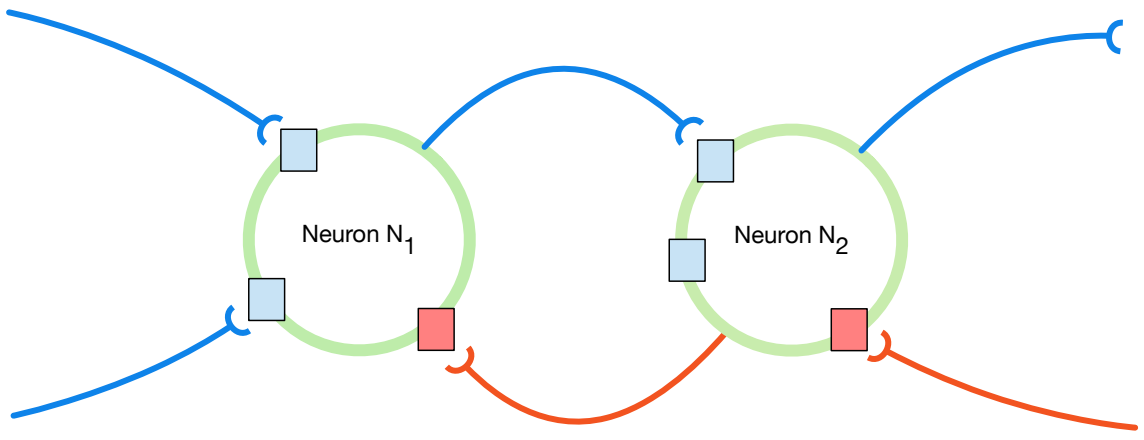

Figure 3: Architecture of neural connection between two BioPPs.

There are three restrictions on BioPP,

- Error signals cannot be calculated using incoming weights because axons convey neural signals unidirectionally.
- A Neuron learns its weights locally based on internal or external error signals.
- All incoming weights should be adaptive.

The neural models proposed in feedback-alignment (FA) and direct feedback-alignment (DFA) are not fully in accordance with the BioPP principles in that some of their incoming weights are fixed. In the following sections, we extend them by optimizing the feedback weights of FA and DFA based on BioPP principles and test the models on benchmark datasets.

# 4 Training BioPP Networks Bidirectionally

Both FA and DFA train neural networks with fixed random weights to propagate error signals[1]. In BioPP, we make those weights adaptive as general incoming weights. For a DNN with 2 hidden layers, the activations in the forward pass are calculated as

$$\overrightarrow{\boldsymbol{a}}_1 = \overrightarrow{\boldsymbol{w}}_1 \boldsymbol{x}, \overrightarrow{\boldsymbol{h}}_1 = \sigma(\overrightarrow{\boldsymbol{a}}_1) \tag{12}$$

$$\overrightarrow{\boldsymbol{a}}_2 = \overrightarrow{\boldsymbol{w}}_2 \overrightarrow{\boldsymbol{h}}_1, \overrightarrow{\boldsymbol{h}}_2 = \sigma(\overrightarrow{\boldsymbol{a}}_2) \tag{13}$$

$$\boldsymbol{a}_y = \overrightarrow{\boldsymbol{w}}_3 \overrightarrow{\boldsymbol{h}}_2, \hat{y} = \sigma_y(\boldsymbol{a}_y) \tag{14}$$

In this section, we propose bidirectional-FA (BFA) and bidirectional-DFA (BDFA) and describe their training pipeline. Then we provide preliminary analysis on why the adaptive feedback weights perform better than fixed feedback weights.

## 4.1 Bidirectional-FA

DNNs with BFA or BDFA learn two mappings between the input and the target output in a two-way manner. To learn these two mappings, we define two loss functions: feedforward loss $\overrightarrow{l}$ and feedback loss $\overleftarrow{l}$, which measure the error in predicting labels and features in the hidden layers or inputs, respectively.

For BFA, the loss functions are:

$$\overrightarrow{l} = -\boldsymbol{y} \cdot \log \hat{\boldsymbol{y}} - (1 - \boldsymbol{y}) \cdot \log(1 - \hat{\boldsymbol{y}}) \tag{15}$$

$$\overleftarrow{l} = \frac{1}{2} \|\hat{\boldsymbol{x}} - \boldsymbol{x}\|_2^2 \tag{16}$$

where $\hat{\boldsymbol{y}}$ and $\hat{\boldsymbol{x}}$ are predicted output and predicted inputs. $\boldsymbol{y}$ and $\boldsymbol{x}$ are target output and target inputs.

We define the forward weights as $\overrightarrow{\boldsymbol{W}}$ and the feedback weights as $\overleftarrow{\boldsymbol{W}}$. The training pipeline includes forward learning phase and backward learning phase, and process them iteratively in each batch.

The gradient at the output layer is calculated as

$$\delta \boldsymbol{a}_y = \frac{\partial \overrightarrow{l}}{\partial \boldsymbol{a}_y} = \hat{\boldsymbol{y}} - \boldsymbol{y} = \overrightarrow{\boldsymbol{e}} \tag{17}$$

For BFA, the gradients for hidden layers in the forward pass are calculated as

$$\delta \overrightarrow{\boldsymbol{a}}_2 = \frac{\partial \overrightarrow{l}}{\partial \overrightarrow{\boldsymbol{a}}_2} = (\overleftarrow{\boldsymbol{w}}_1 \overrightarrow{\boldsymbol{e}}) \odot \sigma'(\overrightarrow{\boldsymbol{a}}_2) \tag{18}$$

$$\delta \overrightarrow{\boldsymbol{a}}_1 = \frac{\partial \overrightarrow{l}}{\partial \overrightarrow{\boldsymbol{a}}_1} = (\overleftarrow{\boldsymbol{w}}_2 \delta \overrightarrow{\boldsymbol{a}}_2) \odot \sigma'(\overrightarrow{\boldsymbol{a}}_1) \tag{19}$$

where $\overleftarrow{\boldsymbol{w}}_i$ is a trainable feedback weight matrix.

Ignoring the learning rate, the updates for the forward weights are calculated as

$$\delta \overrightarrow{\boldsymbol{w}}_1 = -\delta \overrightarrow{\boldsymbol{a}}_1 \boldsymbol{x}^T \tag{20}$$

$$\delta \overrightarrow{\boldsymbol{w}}_2 = -\delta \overrightarrow{\boldsymbol{a}}_2 \overrightarrow{\boldsymbol{h}}_1^T \tag{21}$$

$$\delta \overrightarrow{\boldsymbol{w}}_3 = -\overrightarrow{\boldsymbol{e}} \overrightarrow{\boldsymbol{h}}_2^T \tag{22}$$

where the error signals are transmitted layer by layer through backward weights.

The activations in the feedback pass are then calculated as

$$\overleftarrow{\boldsymbol{a}}_1 = \overleftarrow{\boldsymbol{w}}_1 \boldsymbol{y}, \overleftarrow{\boldsymbol{h}}_1 = \sigma(\overleftarrow{\boldsymbol{a}}_1) \tag{23}$$

$$\overleftarrow{\boldsymbol{a}}_2 = \overleftarrow{\boldsymbol{w}}_2 \overleftarrow{\boldsymbol{h}}_1, \overleftarrow{\boldsymbol{h}}_2 = \sigma(\overleftarrow{\boldsymbol{a}}_2) \tag{24}$$

$$\boldsymbol{a}_x = \overleftarrow{\boldsymbol{w}}_3 \overleftarrow{\boldsymbol{h}}_2, \hat{x} = \sigma_x(\boldsymbol{a}_x) \tag{25}$$

[1] In [18], the authors propose to perform forward-BP and backward-BP training on the same set of weights.

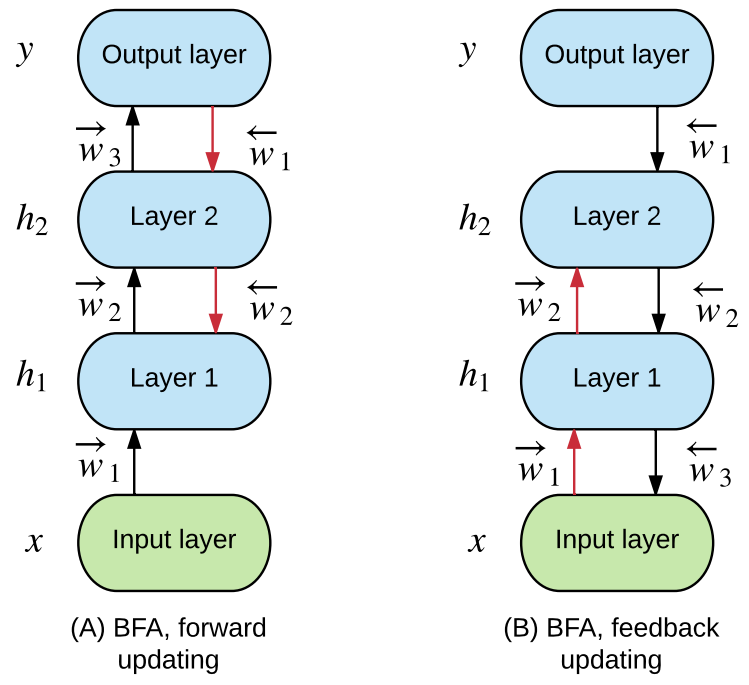

Figure 4: Bidirectional feedback alignment (BFA). Black arrows represent forward activation paths. Red arrows indicate error (gradient) propagation paths. (Modified from [17]).

The gradients for hidden layers in the feedback pass are

$$\delta \boldsymbol{a}_x = \overleftarrow{\boldsymbol{e}} \tag{26}$$

$$\delta \overleftarrow{\boldsymbol{a}}_2 = \frac{\partial \overleftarrow{l}}{\partial \overleftarrow{\boldsymbol{a}}_2} = (\overrightarrow{\boldsymbol{w}}_1 \overleftarrow{\boldsymbol{e}}) \odot \sigma'(\overleftarrow{\boldsymbol{a}}_2) \tag{27}$$

$$\delta \overleftarrow{\boldsymbol{a}}_1 = \frac{\partial \overleftarrow{l}}{\partial \overleftarrow{\boldsymbol{a}}_1} = (\overrightarrow{\boldsymbol{w}}_2 \delta \overleftarrow{\boldsymbol{a}}_2) \odot \sigma'(\overleftarrow{\boldsymbol{a}}_1) \tag{28}$$

where the error signals of backward learning are transmitted through feedforward weights. Ignoring the learning rate, the updates for the feedback weights are calculated as

$$\delta \overleftarrow{\boldsymbol{w}}_1 = -\delta \overleftarrow{\boldsymbol{a}}_1 \boldsymbol{y}^T, \delta \overleftarrow{\boldsymbol{w}}_2 = -\delta \overleftarrow{\boldsymbol{a}}_2 \overleftarrow{\boldsymbol{h}}_1^T, \delta \overleftarrow{\boldsymbol{w}}_3 = -\overleftarrow{\boldsymbol{e}} \overleftarrow{\boldsymbol{h}}_2^T \tag{29}$$

The overall procedure for BFA is shown in Figure 4. The main idea of BFA is that when training one set of weights, the error signals are transmitted layer by layer through the other set of weights. BFA satisfies the principles and restrictions of BioPP. The difference between BFA and target propagation (TP) proposed in [12] is that BFA learns the input features and propagate error signal layer by layer, while each layer in TP learns the output of previous layer with an autoencoder.

## 4.2 Bidirectional-DFA

For BDFA, the loss functions are:

$$\overrightarrow{l} = -\boldsymbol{y} \cdot \log \hat{\boldsymbol{y}} - (1 - \boldsymbol{y}) \cdot \log(1 - \hat{\boldsymbol{y}}) \tag{30}$$

$$\overleftarrow{l}_i = 1 - \sigma(\overleftarrow{\boldsymbol{a}}_i \cdot \overrightarrow{\boldsymbol{h}}_i) \tag{31}$$

where $\hat{\boldsymbol{y}}$ and $\overleftarrow{\boldsymbol{a}}$ are predicted labels and feature maps and $\sigma(x) = \frac{1}{1+e^{-x}}$.

The feedforward and feedback weights are also defined as $\overrightarrow{\boldsymbol{W}}$ and $\overleftarrow{\boldsymbol{W}}$. The training pipeline includes forward learning phase and backward learning phase, and process them iteratively on each training batch. For a DNN with 2 hidden layers, the activations in the forward pass are then calculated as

$$\overrightarrow{\boldsymbol{a}}_1 = \overrightarrow{\boldsymbol{w}}_1 \boldsymbol{x}, \overrightarrow{\boldsymbol{h}}_1 = \sigma(\overrightarrow{\boldsymbol{a}}_1) \tag{32}$$

$$\overrightarrow{\boldsymbol{a}}_2 = \overrightarrow{\boldsymbol{w}}_2 \overrightarrow{\boldsymbol{h}}_1, \overrightarrow{\boldsymbol{h}}_2 = \sigma(\overrightarrow{\boldsymbol{a}}_2) \tag{33}$$

$$\boldsymbol{a}_y = \overrightarrow{\boldsymbol{w}}_3 \overrightarrow{\boldsymbol{h}}_2, \hat{y} = \sigma_y(\boldsymbol{a}_y) \tag{34}$$

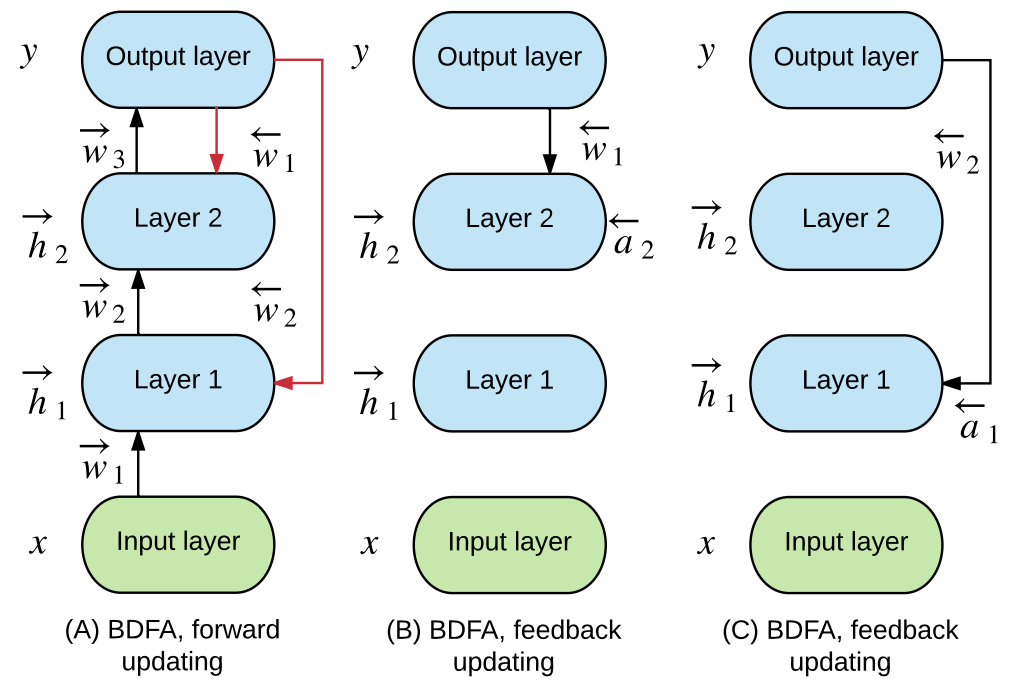

Figure 5: Didirectional direct feedback alignment (BDFA). Black arrows represent forward activation paths. Red arrows indicate error (gradient) propagation paths. (Modified from [17]).

The gradient at the output layer is calculated as

$$\delta \boldsymbol{a}_y = \frac{\partial \overrightarrow{l}}{\partial \boldsymbol{a}_y} = \hat{\boldsymbol{y}} - \boldsymbol{y} = \overrightarrow{\boldsymbol{e}} \tag{35}$$

For BDFA, the gradients for hidden layers in the forward pass are calculated as

$$\delta \overrightarrow{\boldsymbol{a}}_2 = \frac{\partial \overrightarrow{l}}{\partial \overrightarrow{\boldsymbol{a}}_2} = (\overleftarrow{\boldsymbol{w}}_1 \overrightarrow{\boldsymbol{e}}) \odot \sigma'(\overrightarrow{\boldsymbol{a}}_2) \tag{36}$$

$$\delta \overrightarrow{\boldsymbol{a}}_1 = \frac{\partial \overrightarrow{l}}{\partial \overrightarrow{\boldsymbol{a}}_1} = (\overleftarrow{\boldsymbol{w}}_2 \overrightarrow{\boldsymbol{e}}) \odot \sigma'(\overrightarrow{\boldsymbol{a}}_1) \tag{37}$$

where $\overleftarrow{\boldsymbol{w}}_i$ is a trainable feedback weight matrix.

The updates for the forward weights are calculated as

$$\delta \overrightarrow{\boldsymbol{w}}_1 = -\delta \overrightarrow{\boldsymbol{a}}_1 \boldsymbol{x}^T, \delta \overrightarrow{\boldsymbol{w}}_2 = -\delta \overrightarrow{\boldsymbol{a}}_2 \overrightarrow{\boldsymbol{h}}_1^T, \delta \overrightarrow{\boldsymbol{w}}_3 = -\overrightarrow{\boldsymbol{e}} \overrightarrow{\boldsymbol{h}}_2^T \tag{38}$$

In the feedback pass, the activations in the feedback pass are then calculated as

$$\overleftarrow{\boldsymbol{a}}_1 = \overleftarrow{\boldsymbol{w}}_1 \boldsymbol{y}, \overleftarrow{\boldsymbol{a}}_2 = \overleftarrow{\boldsymbol{w}}_2 \boldsymbol{y} \tag{39}$$

The losses on hidden layers are

$$\overleftarrow{l}_1 = 1 - \sigma(\hat{\overleftarrow{\boldsymbol{a}}}_1 \cdot \overrightarrow{\boldsymbol{h}}_1) \tag{40}$$

$$\overleftarrow{l}_2 = 1 - \sigma(\hat{\overleftarrow{\boldsymbol{a}}}_2 \cdot \overrightarrow{\boldsymbol{h}}_2) \tag{41}$$

For BDFA, ignoring the learning rate, the updates for the feedback weights are calculated as

$$\delta \overleftarrow{\boldsymbol{w}}_1 = -\overleftarrow{l}_1 \boldsymbol{y}^T, \delta \overleftarrow{\boldsymbol{w}}_2 = -\overleftarrow{l}_2 \boldsymbol{y}^T \tag{42}$$

The overall procedure for BDFA is shown in Figure 5. The main idea of BDFA model is that each hidden layer calculates a loss separately and updates corresponding feedback weight matrix connecting the hidden layer and output layer.

### 4.3 Approximating BP Gradients with Adaptive Feedback Weights

In this section, we provide analysis on why the adaptive feedback weights applied in bidirectional training models are in principle better than fixed feedback weights. We prove that the overall training performance can be improved if the feedback weights can learn the mapping from output features to input features better.

In [14] the authors prove that random feedback weights act like the pseudoinverse of feedforward weights in the same layer. Following the proof in [14], we consider a linear network with one hidden layer,

$$\boldsymbol{h} = \boldsymbol{A}\boldsymbol{x} \tag{43}$$

$$\hat{\boldsymbol{y}} = \boldsymbol{W}\boldsymbol{h}, \tag{44}$$

where $\boldsymbol{x}$ is input, $\hat{\boldsymbol{y}}$ is output of the network, $\boldsymbol{A} and \boldsymbol{W}$ are feedforward weights. The feedback weight matrix transmitting error signal from output layer to hidden layer is $\boldsymbol{B}$.

Theorem 2 in [14] describes that in FA, the pseudogradients $\delta_{FA}\boldsymbol{h}$ calculated by the random feedback weights satisfy

$$\delta_{FA}\boldsymbol{h} = s\delta_{BP}\boldsymbol{h} \tag{45}$$

where $s$ is a positive scalar, and $\delta_{BP}\boldsymbol{h}$ is the exact gradients calculated by transpose of feedforward weights. As shown in Equation (75) in [14],

$$\delta_{FA}\boldsymbol{h} = \eta(1 - s_y)\boldsymbol{B}\boldsymbol{y} \tag{46}$$

and

$$\delta_{BP}\boldsymbol{h} = \eta(1 - s_y)\boldsymbol{W}^{+}\boldsymbol{y} \tag{47}$$

where $\eta$ and $s_y$ are scalars, $\boldsymbol{B}$ is the random feedback matrix, $\boldsymbol{W}^{+}$ is the pseudoinverse of feedforward matrix $\boldsymbol{W}$ and $\boldsymbol{y}$ is the target output.

Feedback weights are trained by mapping output features to input features. In bidirectional training models, we approximate $\boldsymbol{B}\boldsymbol{y}$ to the hidden layer outputs $\boldsymbol{h}$, that is

$$\boldsymbol{B}\boldsymbol{y} \rightarrow \boldsymbol{h}. \tag{48}$$

As the model is converging, we have

$$\boldsymbol{h} = \boldsymbol{W}^{+}\hat{\boldsymbol{y}} \rightarrow \boldsymbol{W}^{+}\boldsymbol{y}. \tag{49}$$

Thus,

$$\delta_{FA}\boldsymbol{h} \rightarrow \delta_{BP}\boldsymbol{h} \tag{50}$$

We can see that as $\boldsymbol{B}\boldsymbol{y}$ converges to $h$, the gradient calculated with feedback weights will approximate the gradient calculated with transpose of feedforward weights. If backward weights learns the mapping from output features to input features better, then $\delta_{FA}\boldsymbol{h}$ and $\delta_{BP}\boldsymbol{h}$ will be more similar. This gives one explanation to why the adaptive feedback weights outperform the fixed feedback weights. It is worth noting that if backward weights fail to learn the mapping, it might disturb the training of feedforward weights and the convergence of the network.

## 5 Experiments and Discussions

In this section, we investigate if BFA and BDFA can outperform FA and DFA on benchmark datasets with various hyperparameter settings.

We train MLPs on MNIST and CIFAR-10 dataset. The activation functions for hidden layers are Tanh. In order to make the training more stable, the learning rates in all experiments are fixed and set to 0.0001. All the models are trained for 300 epochs. All the results are based on 5 independent runs. The mini-batch size is set to 128. For both MNIST and CIFAR-10 dataset, we use 50,000 samples for training and 10,000 samples for testing.

| Model | BP | FA | DFA | BFA | BDFA |
|---|---|---|---|---|---|
| 1 × 400 | 1.95 | 3.75 | 3.75 | **2.90** | 3.24 |
| 1 × 800 | 1.92 | 4.49 | 4.49 | **3.14** | 3.43 |
| 2 × 400 | 1.71 | 4.25 | 3.70 | **2.89** | 3.28 |
| 2 × 800 | 1.74 | 4.53 | 4.21 | **2.92** | 3.40 |
| 3 × 400 | 1.83 | 5.46 | 3.62 | **2.84** | 3.33 |
| 3 × 800 | 1.80 | 5.48 | 4.30 | **2.95** | 3.37 |

Table 1: Test error rate (%) for back-propagation (BP), feedback alignment (FA), direct feedback alignment (DFA), bidirectional feedback alignment (BFA), and bidirectional direct feedback alignment (BDFA) on MNIST. 1 × 400 indicates that the MLP has 1 hidden layer, each having 400 neurons.

The experimental results on MNIST dataset are summarized in Table 1. We can observe that BFA model performs best on MNIST, and BDFA model outperforms both FA and DFA. Our explanation on the fact that BFA performs better than BDFA is that BFA learns the mapping from output features to input features with a MLP, which has better fitting ability in the backward learning.

To demonstrate the ability of BFA to learn input features, we test if the network can generate input images given output features. For MNIST, the output features are 10-dimension one-hot vectors, according to the classification of inputs. For example, the output feature of digit "4" is $[0, 0, 0, 0, 1, 0, 0, 0, 0, 0]$. Given output features, the network generates the input features by the backward generating procedure described in Equations (23) to (25).

Given the output features of digits "0" to "9", the input features of the digits generated by BFA are shown in Figure 6. The generated images indicate that the feedback weights successfully learn the mapping from the output features to the input features.

The experimental results on CIFAR-10 dataset are summarized in Table 2. Though BFA still outperforms FA and DFA, BDFA has the best performance among all the asymmetric methods on CIFAR-10. One possible reason is that the images of CIFAR-10 are more complicated than those of MNIST dataset, which makes it difficult for BFA-based network to learn the input features. According to our proof in section 4.3, the performance of feedforward weights will be compromised if the backward learning process fails to map the output features to the input features. The better performance of BDFA in this case mcan be attributed to the fact that it is only required to learn the features in the hidden layers, which is easier compared to the task to learn the raw inputs directly. In other words, those features captured in the hidden layers are more abstract and in a low-dimensional space. However, MLPs are not good at mapping the output features to the raw input images, as the features of which are not abstract enough. BFA model forces the network to fit input images with backward weights, and the numerical stability in training is seriously compromised. DFA faces the similar issue when learning convolutional weights on CIFAR-10 [17], which tried to learn transmitting error signal from output layer directly to convolutional layers. In future work, we plan on proposing more stable models that can convey the backward teaching signals to the layers which can produce more complicated outputs.

To better learn the backward features on CIFAR-10, we slightly modify the backward training in BDFA. The output features used to train the feedback weights on CIFAR-10 dataset are now calculated as:

$$\boldsymbol{y}' = \boldsymbol{y} + \alpha\hat{\boldsymbol{y}} \tag{51}$$

where $\boldsymbol{y}'$ is the output features we actually used in training BDFA model on CIFAR-10 dataset, $\boldsymbol{y}$ is the target output features, $\hat{\boldsymbol{y}}$ is output of the network calculated with current parameters, and $\alpha$ is a small positive scalar. In our experiments, $\alpha$ is set to 0.25. We apply this modification because the current outputs contain certain amount of random features of input images. Given these random features, the negative effect of randomness of the input features is mitigated and thus it is easier for backward weights to learn the mapping from output features to hidden layer outputs and input features.

BFA and BDFA demonstrate novel applications of the adaptive asymmetric gradient-based methods for optimizing DNNs. Especially in BDFA, the learning of the feedback weights and the learning of the feed forward weights are disconnected in the sense that the feedback weights are unaware of the existence of the feed forward weights. This learning process for feedback updating of BDFA is also consistent with the insight [11, 19, 9] that errors can result from mismatches between the actual and the expected perceptions, rather than coming from external teaching signals.

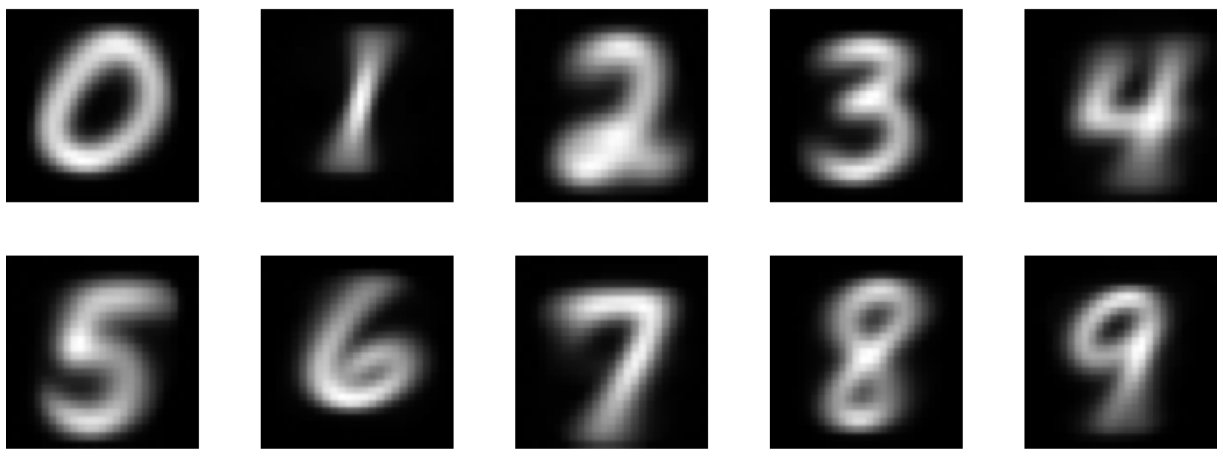

Figure 6: Feature maps learned by BFA-trained MLP.

| Model | BP | FA | DFA | BFA | BDFA |
|---|---|---|---|---|---|
| 1 × 400 | 46.78 | 48.38 | 48.48 | 48.34 | **48.00** |
| 1 × 800 | 45.22 | 48.08 | 48.04 | 48.01 | **47.36** |
| 2 × 400 | 46.22 | 49.50 | 48.52 | 48.38 | **47.98** |
| 2 × 800 | 44.97 | 48.88 | 48.86 | 48.55 | **48.46** |

Table 2: Test error rate (%) for back-propagation (BP), feedback alignment (FA), direct feedback alignment (DFA), bidirectional feedback alignment (BFA), and bidirectional direct feedback alignment (BDFA) on CIFAR-10. $1 \times 400$ indicates that the MLP has 1 hidden layer, each having 400 neurons.

## 6 Conclusion

In this work, we proposed biologically plausible perceptron paradigm based on related literature in neuroscience. We also designed and evaluated Bidirection-FA and Bidirectional-DFA models on benchmark datasets. To the best of our knowledge, this is the first research attempt to show that adaptive asymmetric feedback channels are more effective than random and fixed feedback channels in DNNs. Although it is not clear if the brain implements this particular form of adaptive feedback, it is a step towards better understanding how the brain supports learning from error signals.

## References

[1] Atilim Gunes Baydin, Barak A Pearlmutter, and Alexey Andreyevich Radul. Automatic differentiation in machine learning: a survey. *arXiv preprint arXiv:1502.05767*, 2015.

[2] Yoshua Bengio, Dong-Hyun Lee, Jorg Bornschein, and Zhouhan Lin. Towards biologically plausible deep learning. *arXiv preprint arXiv:1502.04156*, 2015.

[3] Yoshua Bengio, Dong-Hyun Lee, Jorg Bornschein, Thomas Mesnard, and Zhouhan Lin. Towards biologically plausible deep learning. *arXiv preprint arXiv:1502.04156*, 2015.

[4] Timothy V Bliss and Graham L Collingridge. A synaptic model of memory: long-term potentiation in the hippocampus. *Nature*, 361(6407):31, 1993.

[5] John D Clements, Robin AJ Lester, Gang Tong, Craig E Jahr, and Gary L Westbrook. The time course of glutamate in the synaptic cleft. *SCIENCE-NEW YORK THEN WASHINGTON-*, 258:1498–1498, 1992.

[6] Peter Dayan and Laurence F Abbott. *Theoretical neuroscience*, volume 806. Cambridge, MA: MIT Press, 2001.

[7] Dumitru Erhan, Yoshua Bengio, Aaron Courville, Pierre-Antoine Manzagol, Pascal Vincent, and Samy Bengio. Why does unsupervised pre-training help deep learning? *Journal of Machine Learning Research*, 11(Feb):625–660, 2010.

[8] M Foster. A textbook of physiology. part iii, 1897.

[9] Karl Friston. The free-energy principle: a unified brain theory? *Nature Reviews Neuroscience*, 11(2):127–138, 2010.

[10] Martin T Hagan. *Neural network design*, volume 20.

[11] Geoffrey Hinton. The ups and downs of hebb synapses. *Canadian Psychology/Psychologie canadienne*, 44(1):10, 2003.

[12] Dong-Hyun Lee, Saizheng Zhang, Asja Fischer, and Yoshua Bengio. Difference target propagation. In *Joint European Conference on Machine Learning and Knowledge Discovery in Databases*, pages 498–515. Springer, 2015.

[13] Irwin B Levitan and Leonard K Kaczmarek. *The neuron: cell and molecular biology*. Oxford University Press, USA, 2015.

[14] Timothy P Lillicrap, Daniel Cownden, Douglas B Tweed, and Colin J Akerman. Random feedback weights support learning in deep neural networks. *arXiv preprint arXiv:1411.0247*, 2014.

[15] Timothy P Lillicrap, Daniel Cownden, Douglas B Tweed, and Colin J Akerman. Random synaptic feedback weights support error backpropagation for deep learning. *Nature Communications*, 7, 2016.

[16] MA Lynch. Long-term potentiation and memory. *Physiological reviews*, 84(1):87–136, 2004.

[17] Arild Nøkland. Direct feedback alignment provides learning in deep neural networks. *arXiv preprint arXiv:1609.01596*, 2016.

[18] Adigun Olaoluwa and Kosko Bart. Bidirectional representation and backpropagation learning. *International Conference on Advances in Big Data Analytics*, 2016.

[19] Randall C O'Reilly. Biologically plausible error-driven learning using local activation differences: The generalized recirculation algorithm. *Neural computation*, 8(5):895–938, 1996.

[20] David E Rumelhart, Geoffrey E Hinton, and Ronald J Williams. Learning representations by back-propagating errors. *Cognitive modeling*, 5(3):1, 1988.

[21] Wikipedia. Long-term potentiation — Wikipedia, the free encyclopedia. http://en.wikipedia.org/w/index.php?title=Long-term%20potentiation&oldid=765527653, 2017. [Online; accessed 22-February-2017].

[22] Sam Wiseman, Sumit Chopra, Marc'Aurelio Ranzato, Arthur Szlam, Ruoyu Sun, Soumith Chintala, and Nicolas Vasilache. Training language models using target-propagation. *arXiv preprint arXiv:1702.04770*, 2017.